\pdfoutput=1
\documentclass[11pt]{article}

\usepackage[final]{acl}

\usepackage{times}
\usepackage{latexsym}
\usepackage{makecell}
\usepackage{array}
\usepackage[T1]{fontenc}
\usepackage[utf8]{inputenc}
\usepackage{microtype}
\usepackage{booktabs}
\usepackage{inconsolata}
\usepackage{graphicx}
\usepackage{multirow}
\usepackage{tabularx}
\usepackage{balance}
\usepackage{dblfloatfix}

\title{WisdomInterrogatory (LuWen): An Open-Source Legal Large Language Model Technical Report}

\author{
  Yiquan Wu, Yuhang Liu, Yifei Liu, Ang Li, Siying Zhou, Kun Kuang\thanks{Corresponding author.}, Fei Wu \\
  Zhejiang University, Hangzhou, China \\
  \texttt{\{wuyiquan, kuangkun\}@zju.edu.cn}
}

\begin{document}
\maketitle

\begin{abstract}
Large language models have demonstrated remarkable capabilities across a wide range of natural language processing tasks, yet their application in the legal domain remains challenging due to the specialized terminology, complex reasoning requirements, and rapidly evolving legal knowledge involved. In this paper, we present WisdomInterrogatory (LuWen), an open-source Chinese legal language model built upon the Baichuan foundation model through three key techniques: continual pre-training on a large-scale legal corpus, supervised fine-tuning with carefully curated legal instruction data, and retrieval-augmented generation integrated with a comprehensive legal knowledge base. We evaluate LuWen on five representative legal tasks spanning both prediction and generation settings, including legal judgment prediction, judicial examination, legal text summarization, law article question answering, and judicial decision reasoning. Experimental results show that LuWen outperforms several strong baselines, demonstrating the effectiveness of our approach in adapting general-purpose language models to the legal domain.
\end{abstract}

\section{Introduction}

Large language models (LLMs), especially those based on the Transformer \citep{vaswani2017attention} architecture such as ChatGPT and Gemini, utilize deep learning techniques to excel in the comprehension and generation of human language, often containing billions or even hundreds of billions of parameters. By pre-training on vast amounts of text data collected from the internet, these models acquire capabilities in recognizing complex language patterns, understanding sentence structure, and grasping contextual meanings \citep{wu2023brief}. They perform tasks such as text generation, reading comprehension, dialogue, translation, and summarization, and have found widespread use in applications like automatic article writing, code generation, and customer support \citep{chang2023survey}. By comprehending context and generating coherent text, these models provide powerful support across fields such as education and media. As a significant branch of natural language processing, they also demonstrate vast potential in processing legal texts, offering promising applications for courts, law firms, academic institutions, and research. However, applying these general-purpose models in the legal domain presents a range of unique challenges.

First, legal texts contain extensive specialized terminology and highly structured language, requiring models not only to understand standard language but also to grasp domain-specific legal knowledge. Second, legal tasks are explicitly defined and often based on complex legal principles and precedents that require clear reasoning, demanding enhanced task comprehension and logical inference capabilities from the models. Furthermore, continuous updating of legal rules and knowledge challenges models in terms of accuracy and timeliness of information. These challenges illustrate that, despite the broad success of large models across various domains, specific optimizations and adaptations are necessary for their effective application in legal contexts.

To address these challenges, we have developed an open-source Chinese legal language model, WisdomInterrogatory (LuWen)\footnote{\url{https://github.com/zhihaiLLM/wisdomInterrogatory}}, built on the foundation of a general large language model. Our technical approach consists of three main components:

\begin{itemize}
    \item \textbf{Continual Pre-Training (CPT)}: We perform CPT on a specialized legal corpus to enhance the model’s ability to understand legal-specific language patterns and knowledge structures.
    \item \textbf{Supervised Fine-Tuning (SFT)}: We employ SFT, training the model on specific legal tasks to enable it to understand distinct legal requirements and provide more precise and reasoned responses—significantly improving the model's performance in targeted legal applications.
    \item \textbf{Retrieval Augmented Generation (RAG)}: Finally, we utilize the RAG method to link the model with legal databases, enabling it to access and reference continually updated legal information and case law. This approach ensures that even amidst rapid changes in the legal knowledge landscape, the model maintains accuracy and timeliness in its information output.
\end{itemize}

In summary, this technical approach not only enhances the model’s expertise in the legal domain but also ensures the real-time relevance and reliability of its outputs.

\section{Method}

\begin{figure*}[!t]
\centering
\includegraphics[width=\textwidth]{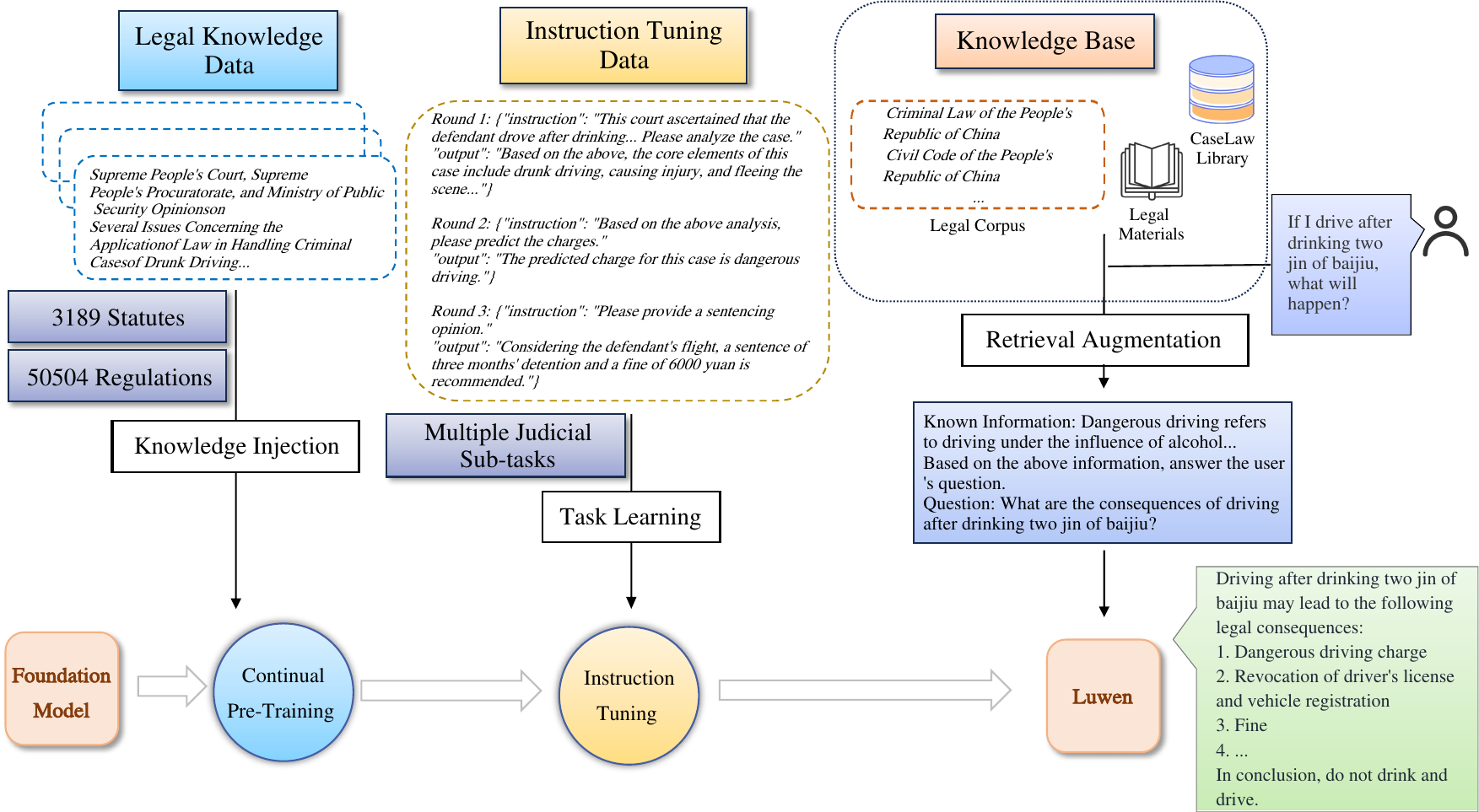}
\caption{
LuWen Technology Roadmap.
}
\label{fig:method}
\end{figure*}

As shown in Figure~\ref{fig:method}, the technical approach of the LuWen model is divided into three main components: continual pre-training, instruction fine-tuning/supervised fine-tuning, and retrieval-augmented generation. This section will elaborate on each of these components.

\subsection{Continual Pre-Training, CPT}
In the legal domain, where data are characterized by high specialization and complex semantics, foundational models trained on general datasets often fall short of achieving optimal performance. This limitation stems from the model's insufficient capabilities in semantic understanding and language generation within specialized legal contexts. To address this issue, we implemented a strategy of continual pre-training to enhance the model's comprehension and handling of professional legal texts through further training.

The continual pre-training process is technically similar to the initial pre-training, based on a self-supervised learning framework. The model maximizes the conditional probability to predict the next word following a given preceding sequence by optimizing the following objective function to adjust the model parameters, $\theta$:

\begin{equation}
\theta = \arg \max_\theta \sum_{t=1}^T \log p(w_t | x_{<t}; \theta)
\end{equation}

where $\theta$ denotes the model parameters, $T$ represents the length of the text sequence, $x_{<t}$ is the preceding word sequence, and $w_t$ is the target word following the sequence.

In our implementation, we first compile a large Chinese legal corpus encompassing various types of texts, such as legal books, academic papers, statutes, court rulings and contracts, sourced from legal databases, judicial documents and law journals. To ensure the quality of the corpus, we completed an initial filtering and cleaning by removing duplicate, erroneous, and irrelevant data.

For LuWen, we chose the Baichuan1-7B \citep{baichuan} model as the foundation for continual pre-training due to its outstanding capabilities in handling Chinese. We conducted the continual pre-training using a comprehensive corpus containing 200 GB of data, with 20\% comprising Chinese legal texts and 80\% general texts. This ratio was chosen to ensure that while the model acquires legal knowledge, it also retains general capabilities. It is noteworthy that, in continual pre-training, all data are shuffled and concatenated directly without reformatting into a specific data structure.

\subsection{Supervised Fine-Tuning, SFT}

The supervised fine-tuning process, also referred to as instruction tuning, aims to train the model to more accurately comprehend human intent in input prompts, thereby enhancing its capability to perform specific tasks. Compared to the prior CPT phase, the data used in the instruction tuning stage adopts a more standardized format, specifically in the form of instruction-response pairs. Each instruction is prefixed with 'Human:', and each response with 'Assistant:', with the goal of guiding the model more effectively into a conversational mode rather than simply remaining in a text completion phase.

\begin{table*}[!t]
\centering
\small
\caption{Composition of SFT Data Used in Model Training}
\begin{tabular}{
  >{\arraybackslash}m{0.25\textwidth}
  >{\arraybackslash}p{0.50\textwidth}
  >{\centering\arraybackslash}m{0.06\textwidth}
  >{\centering\arraybackslash}m{0.06\textwidth}
}
\toprule % First horizontal line
\textbf{Dataset} & 
\textbf{Data Description} & 
\textbf{Volume} & 
\textbf{Source} \\
\midrule % Second horizontal line
\multicolumn{4}{c}{\textit{Legal Domain}} \\ % Merge four columns and center "Legal Domain"
\midrule % Third horizontal line
Legal Consultation & Q\&A in legal consultation contexts & 7k & \makecell{Public/\\Private} \\ 
Legal Examination & Q\&A from legal examination settings & 6k & \makecell{Private} \\ 
Scenario-based Legal Q\&A & Legal Q\&A based on specific scenarios & 4k & \makecell{Public} \\ 
Legal Reading Comprehension & Q\&A tasks for legal reading comprehension & 4k & \makecell{Private} \\
Case Summary Extraction & Extraction of abstract information from legal cases & 3k & \makecell{Private} \\
Court Opinion Generation & Generation of court opinions based on case data & 3k & \makecell{Private} \\
Multi-round Legal Consultation & Multi-round Q\&A in legal consultations & 0.8k & \makecell{Private} \\
Charge Prediction & Case data with corresponding charges & 0.5k & \makecell{Private} \\
Legal Provision Prediction & Case data with corresponding violated provisions & 0.5k & \makecell{Private} \\
Judgment Prediction & Case data with corresponding final judgments & 0.5k & \makecell{Private} \\
Monetary Amount Extraction & Case data with total involved amounts & 0.2k & \makecell{Private} \\
Others & Miscellaneous legal data such as document templates & 0.5k & \makecell{Private} \\

\textbf{Total} &  & \textbf{30k} & \\
\midrule % Horizontal line
\multicolumn{4}{c}{\textit{General Domain}}\\
\midrule % Horizontal line
Open-Orca (GPT-4) & Stratified sample from the Open-Orca GPT-4 dataset & 20k & \makecell{Public} \\
MOSS (Chinese Section) & Multi-round dialogue data from bilingual conversations & 15k & \makecell{Public} \\
ShareGPT & Conversations including user inputs and ChatGPT responses & 12.5k & \makecell{Public} \\
Belle & Chinese instruction data covering 13 categories & 10k & \makecell{Public} \\
C3 & Chinese reading comprehension dataset with multiple-choice questions & 6k & \makecell{Public} \\
Puffin & Multi-round dialogues with GPT-4 and fine-tuned responses & 3k & \makecell{Public} \\
School Math & Chinese math problems with solutions & 2k & \makecell{Public} \\
LIMA & Dataset with curated prompts and responses & 1k & \makecell{Public} \\
Self-awareness \& Greetings & Model's preset information and greetings & 0.5k & \makecell{Private} \\

\textbf{Total} & & \textbf{70k} & \\
\bottomrule
\end{tabular}
\label{tab:data}
\end{table*}

Research has consistently shown that during the instruction tuning phase, the quality of data is significantly more important than the quantity. Following this principle, we have rigorously controlled the scale of the instruction dataset. Specifically, as shown in Table~\ref{tab:data}, we used a total of 100,000 instruction tuning data points, with data from the legal domain accounting for 30\%. This legal data primarily covers areas such as judicial consultation, legal scenario-based Q\&A, legal violations and charge prediction, sentencing prediction, legal examination questions, court opinions, case summary extraction, and legal reading comprehension. The remaining 70\% consists of general-purpose data, including various multilingual datasets in Chinese and English, covering a range of natural language processing tasks. This distribution is intended to ensure that the model retains general capabilities while learning new, domain-specific instructions.

To obtain the instruction dataset, we sourced it from two primary avenues: publicly available Q\&A datasets and our self-collected legal texts. For Q\&A datasets, we implemented a stringent data filtering process; for legal texts, we converted the narrative format to Q\&A format.

\textbf{Data Filtering.} The primary challenges in data filtering were the vast volume and inconsistent quality of publicly available Q\&A datasets. To ensure high-quality Q\&A data, the filtering process was conducted from both the instruction and response perspectives.

From the instruction perspective, we focused on three core elements: balancing capabilities, sample diversity, and conversational complexity. To achieve capability balance—especially when handling large datasets like the legal consultation dataset—we applied under-sampling to reduce its proportion in the dataset. Sample diversity was achieved by selecting representative samples within each dataset. We utilized clustering algorithms such as K-Means to categorize the dataset into multiple classes, then performed random sampling within each class to ensure broad coverage. Additionally, to enhance the model's reasoning ability, we retained multi-turn conversation data, as these often showcase complex logical transitions.

From the response perspective, we first constructed a sensitive word dictionary to filter out responses that might contain harmful information. Given the constant updates to legal information, we adapted to changes in legal provisions through content replacement. For example, with the implementation of the PRC Civil Code, many old legal terms were updated, so we built a mapping dictionary to convert old legal terms to their new usages. Furthermore, to improve user interaction, we included Q\&A that featured self-awareness, allowing the model to briefly introduce itself and help users better understand its application scenarios.

\begin{figure*}[!t]
  \centering
  \includegraphics[width=0.8\textwidth]{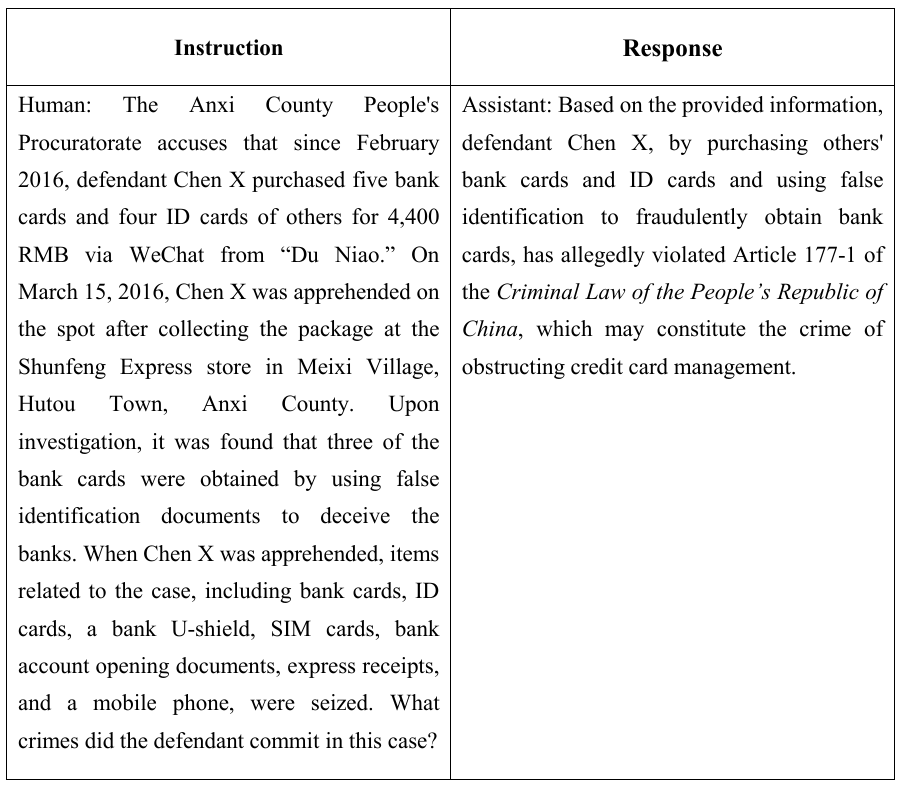}
  \caption{An Example of Constructed Instruction-Response Data}
  \label{fig2}
\end{figure*}

\textbf{Data Construction.} In terms of data construction, we primarily dealt with self-collected legal texts. Although these texts are of relatively high quality, they are mostly in narrative form rather than question-answer format. Therefore, we needed to abstract these texts based on task requirements and reconstruct them to generate a data format suitable for instruction tuning. For example, in the task of charge prediction within judgment documents, we first extracted the statement of facts and the judgment charges. Then, we divided the work into manual and automated construction phases to complete the dataset.

During the manual construction phase, the main tasks included defining representative seed instructions and clarifying response steps. Seed instructions refer to typical prompts that reflect task requirements. For instance, in the charge prediction task, "What charges did the defendant face in this case?" serves as a seed instruction. The response steps were defined to ensure that responses are not merely brief labels but structured explanations. In the charge prediction task, our response steps were: 1. Summarize and abstract the case; 2. List relevant legal provisions; 3. State the charges faced by the defendant.

In the automated construction phase, we used ChatGPT to help expand the seed instructions into over 100 question templates, preserving their original intent to enhance instruction diversity. We then provided ChatGPT with the case facts description, judgment charges, and response steps to guide it in generating high-quality answers.

Finally, by merging case facts with the question templates, we generated a series of instruction prompts. These prompts, along with structured responses, formed our meticulously constructed instruction-response dataset. Examples are shown in Figure~\ref{fig2}.

Throughout the data construction process, we successfully created over ten types of legal tasks covering various dimensions such as legal provision prediction, charge prediction, sentencing prediction, monetary amount extraction, court opinion generation, knowledge base reading, and choice questions for legal examinations. These tasks not only cover multiple key aspects of the legal domain but also significantly enhance the model's ability to handle a wide range of legal-related issues.

\subsection{Retrieval Augmented Generation, RAG}
Retrieval augmented generation is a method that incorporates relevant knowledge before the model provides answers, helping the model respond more accurately. Specifically, this technology transforms the model from a "closed-book" state to an "open-book" state, significantly improving the quality and accuracy of responses. The main advantages of this method include: first, introducing external knowledge to the model effectively reduces the likelihood of generating incorrect information; second, given the rapid evolution of legal knowledge and the high cost of retraining the model to incorporate these updates, connecting to an external knowledge base allows the model to stay current with knowledge updates, which is especially important in the legal field.

In this context, we primarily constructed a legal knowledge base and proposed a multi-source, multi-path legal knowledge retrieval process.

\textbf{Construction of Legal Knowledge Base.} The construction of the legal knowledge base involves several stages, beginning with knowledge collection, which includes:

\begin{itemize}
    \item \textbf{Law Library:} Encompassing the constitution, various types of laws (such as civil and commercial law, criminal law), judicial interpretations, regional regulations, oversight regulations, and administrative laws.
    \item \textbf{Legal Document Template Library:} Covering templates for complaints, appeals, court judgments, and contracts, as well as other commonly used legal documents.
    \item \textbf{Legal Textbook Library:} Including textbooks on civil law, international private law, and environmental resource law, among other fields.
    \item \textbf{Case Library:} Consisting of numerous cases in criminal and civil law.
    \item \textbf{National Unified Legal Professional Qualification Examination Library:} Containing exam questions, options, explanations, and answers from 2016 to 2020.
    \item \textbf{Legal Q\&A Library:} Collecting common legal questions and answers from various fields.
\end{itemize}

The next step involves storing the knowledge. To optimize the retrieval model's performance across different knowledge bases, we standardized the knowledge format. Each piece of knowledge is stored as a key-value pair, where the key is a brief description of the knowledge point to facilitate retrieval, and the value contains the detailed content for the model’s reference. This format not only enables a decoupling between knowledge retrieval and augmentation but also improves the model’s efficiency in matching knowledge and ensures the accuracy of the content.

\begin{figure*}[!t]
  \centering
  \includegraphics[width=\textwidth]{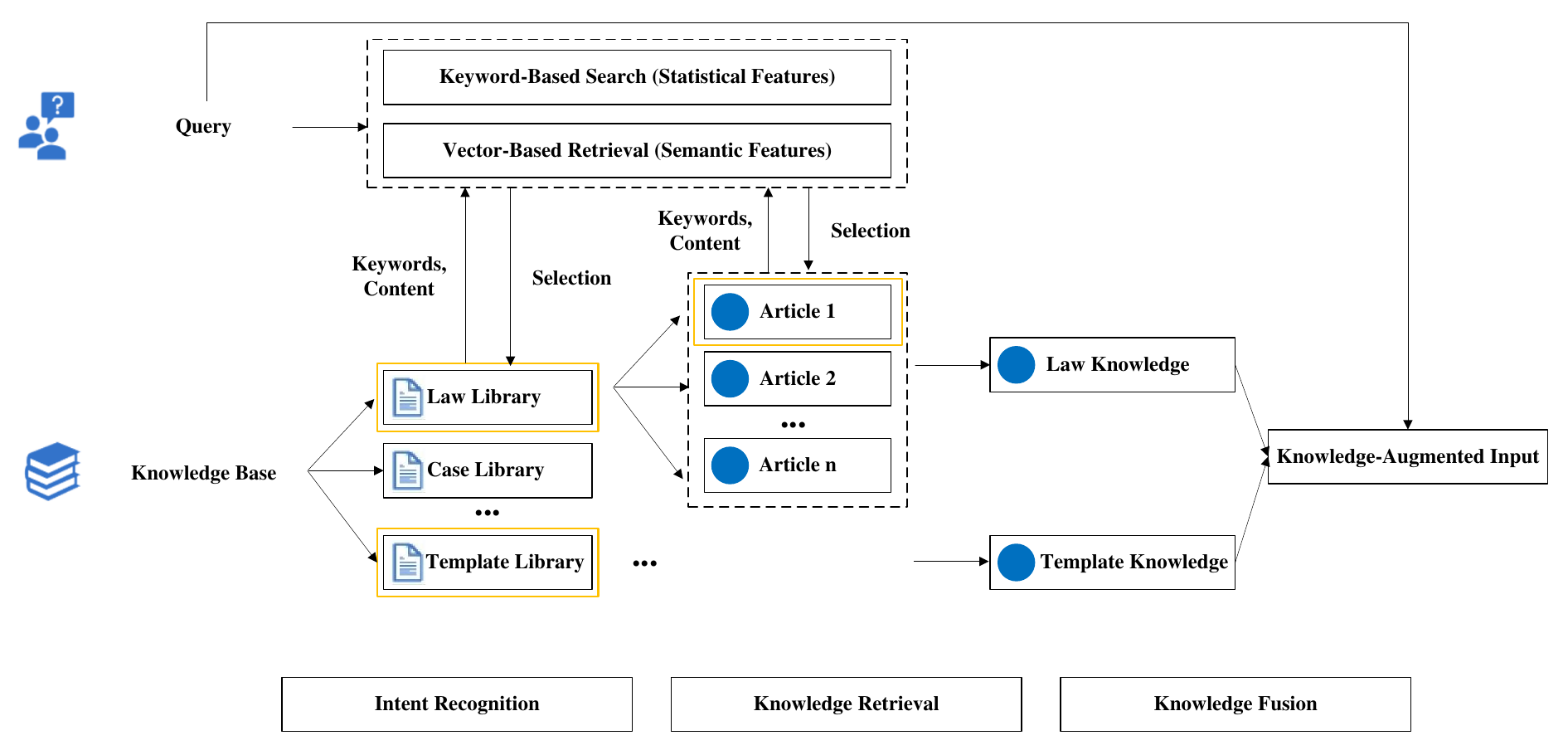}
  \caption{Multi-Source, Multi-Path Legal Knowledge Retrieval}
  \label{fig3}
\end{figure*}

\textbf{Multi-source, Multi-path Legal Knowledge Retrieval.} Legal knowledge retrieval involves retrieving relevant knowledge based on a given instruction. Considering the diversity of legal tasks, as shown in Figure~\ref{fig3}, we propose a multi-source, multi-path legal knowledge retrieval process, which includes the following steps:
\begin{enumerate}
    \item Intent Recognition: Since we have multi-source knowledge bases, not all of them will be relevant to the current instruction. Therefore, the first step is matching at the knowledge base level, known as intent recognition. Specifically, we match keywords in the question with those in the knowledge bases to identify the type of knowledge involved in the question. This enables us to selectively use the appropriate knowledge base to assist, thereby reducing confusion.
    \item Knowledge Retrieval: This process employs both retrieval based on statistical features and retrieval based on semantic features. Statistical feature retrieval focuses on the match between keywords in the knowledge base entries and keywords in the query, while semantic feature retrieval uses a trained retrieval model to calculate the vector similarity between the input query and the knowledge keys. In practice, results from statistical feature retrieval are prioritized, followed by results from semantic feature retrieval. Additionally, we set a similarity threshold for semantic retrieval, returning only results that meet the threshold.
    \item Semantic Retrieval Model Training: For the semantic retrieval model, we conduct additional training using contrastive learning. We select actual legal statutes, similar cases, and relevant book descriptions as positive examples, while negative examples are chosen either randomly or based on high similarity with mismatched content.
    \item Knowledge Fusion: Based on the multi-source knowledge retrieved, we fuse this information and provide it to the large legal model. For instance, when querying the judgment of a specific case, if relevant knowledge in the law library and the case library is retrieved, a specialized instruction containing these knowledge points is constructed and input into the model to assist it in making a more accurate judgment.
\end{enumerate}

This approach of integrating multiple knowledge sources not only improves the model’s response quality but also allows it to quickly adapt to updates in legal knowledge, which is crucial for applications in the legal domain.

\section{Experiment}

\subsection{Introduction to Evaluation Tasks}
As a comprehensive model in the legal domain, LuWen is applicable in a variety of scenarios, making its evaluation a challenging task. Here, we have selected five common tasks to evaluate its performance for reference.

\paragraph{Prediction Tasks}
In prediction tasks, we selected the following two tasks, with their definitions and evaluation metrics introduced as follows:

$\bullet$ \textbf{Legal Judgment Prediction} is one of the most common tasks in Chinese legal proceedings. The input consists of the facts of the crime, and the output includes the legal statutes, charges, and sentences. The test set for this task consists of 2,490 samples, and the evaluation metric is accuracy.

$\bullet$ \textbf{Judicial Examination} is one of the most challenging examinations in China and plays a critical role in the career of legal professionals. Therefore, this task is based on question-answering for the judicial examination. The task requires the model to generate the correct answer given a question and a set of options. The test set for this task consists of 2,108 samples, and the evaluation metric is exact match, where a prediction is considered correct only if the selected option matches the correct option precisely.

\paragraph{Generation Tasks}
In generation tasks, we selected the following three tasks, with their definitions and evaluation metrics introduced as follows:

$\bullet$ \textbf{Legal Text Summarization} involves compressing, condensing, and summarizing the content of judicial documents, reflecting the judicial process, facts, reasoning, and legal grounds. This task requires generating a legal summary text based on the original text of the judicial document. The sample size for the test set is 50, and it uses human evaluation as the metric, focusing on the following criteria:

(1) \textbf{Accuracy $s_1$}: Assesses whether the summary accurately reflects the facts, reasoning, and judicial grounds in the original judicial document. Scores range from 1 to 5, where 1 indicates major errors in most important information, and 5 indicates perfect accuracy.

(2) \textbf{Completeness $s_2$}: Assesses whether the summary comprehensively includes all key information about the case. Scores range from 1 to 5, where 1 represents minimal inclusion of key information, and 5 indicates highly comprehensive information.

(3) \textbf{Readability $s_3$}: Assesses whether the language of the summary is fluent and its logical structure is clear. Scores range from 1 to 5, where 1 indicates the summary is almost unreadable, and 5 denotes excellent fluency and logical clarity.

The calculation formula for the evaluation score is as follows:
\begin{equation}
Q_1 = (s_1 + s_2 + s_3) / (3 \times 5)
\end{equation}
where $s_1$, $s_2$, and $s_3$ are the scores for accuracy, completeness, and readability on a scale of 5.

$\bullet$ \textbf{Law Article Question Answering} involves recommending relevant legal articles and providing pertinent legal interpretations based on legal questions posed by users and descriptions of case facts. It uses human evaluation as the metric, focusing on the following aspects, with a sample size of 50:

(1) \textbf{Relevance of Mentioned Articles $s_4$}: Measures the relevance of recommended legal articles to the user's query, scored from 1 to 5, where 1 denotes almost no relevance, and 5 signifies high relevance.

(2) \textbf{Accuracy in Application of Articles $s_5$}: Evaluates the accuracy of the legal interpretations provided in the context of a specific case, scored from 1 to 5, where 1 indicates a serious mismatch with case facts or legal articles, and 5 denotes full alignment.

The calculation formula for the evaluation score is as follows:
\begin{equation}
Q_2 = (s_4 + s_5) / (2 \times 5)
\end{equation}
where $s_4$ and $s_5$ are the scores for relevance and accuracy of application on a scale of 5.

$\bullet$ \textbf{Judicial Decision Reasoning} involves reasoning and making decisions based on case facts. This task considers facts investigated by the court, contracts signed between the plaintiff and the defendant, the plaintiff's claims, and other evidence. During this process, it is necessary to determine applicable legal grounds, identify contentious focal points, and ultimately provide specific judgment outcomes. The sample size of the test set is 40, and the evaluation criteria include completeness, relevance, and correctness:

(1) \textbf{Completeness $s_6$}: Measures the system’s ability to capture case facts and relevant legal articles, scored from 1 to 5, where 1 indicates capturing only minimal facts and articles, and 5 indicates a comprehensive understanding of all aspects of the case.

(2) \textbf{Relevance $s_7$}: Measures the system’s accuracy in identifying and applying legal articles and grounds, scored from 1 to 5, where 1 indicates identified legal articles are almost irrelevant to the case, and 5 indicates the articles and grounds most relevant to the case are accurately identified.

(3) \textbf{Correctness $s_8$}: Assesses the accuracy of the system’s final judgment, scored from 1 to 5, where 1 means decisions seriously contradict actual legal provisions and precedent judgments, and 5 means full congruence.

The calculation formula for the evaluation score is as follows:
\begin{equation}
Q_3 = (s_6 + s_7 + s_8) / (3 \times 5)
\end{equation}
where $s_6$, $s_7$, and $s_8$ are the scores for completeness, relevance, and correctness on a scale of 5.

\subsection{Baseline Models}

For the prediction tasks, we adopted Baichuan2-Chat \cite{baichuan}, GLM2 \cite{glm}, Qwen \cite{qwen}, and InternLM2 \cite{internlm} as baseline models, all of which are multilingual large language models featuring extended context lengths, superior multi-turn dialogue capabilities, and tool utilization abilities. GPT-3.5 \cite{instructgpt} is a large-scale language model proposed by OpenAI that achieves optimal performance on various natural language tasks. The specific version we used is GPT-3.5-Turbo-0613.

For the generation tasks, we used the following methods as baseline models:

Baichuan+SFT is based on Baichuan with supervised fine-tuning on legal-domain data. GPT-3.5 \cite{instructgpt} serves as a baseline model, the same as that in the prediction tasks.

\begin{table*}[!t]
\caption{Results of Prediction Tasks}
\centering
\small
\begin{tabular}{lcccc}
\toprule
\multirow{2}{*}{\textbf{Method}} & \multicolumn{3}{c}{\textbf{Legal Judgment Prediction}} & \multirow{2}{*}{\textbf{Judicial Examination}} \\
 & \textit{Statute} & \textit{Charge} & \textit{Sentence} &  \\
\midrule
Baichuan2-Chat & 0.15 & 0.00 & 0.20 & 0.00 \\
GLM2 & 0.22 & 0.28 & \textbf{0.44} & 0.11 \\
Qwen & 0.04 & 0.03 & 0.36 & 0.09 \\
InternLM2 & 0.11 & 0.10 & 0.01 & 0.09 \\
GPT-3.5 & 0.22 & 0.12 & 0.42 & \textbf{0.22} \\
\midrule
LuWen & \textbf{0.35} & \textbf{0.31} & 0.41 & 0.18 \\
\bottomrule
\end{tabular}
\label{tab: predict_res}
\end{table*}

\subsection{Evaluation Results}

\paragraph{Results of Prediction Tasks}

From Table~\ref{tab: predict_res}, we have the following observations:
(1) Our model achieves the best results in predicting legal statutes and charges for legal judgment prediction and significantly outperforms most baseline models in other tasks. This result demonstrates that supervised fine-tuning in the legal domain can markedly enhance model performance on legal tasks.
(2) Generative models generally exhibit lower results on prediction tasks, indicating room for improvement compared to traditional prediction models.
(3) Compared to \textbf{Baichuan2}, \textbf{LuWen} shows better performance across all prediction tasks, which validates the effectiveness of continual pre-training on the Baichuan base model.
(4) Additionally, \textbf{Baichuan2} scores lower in charge prediction and judicial examination because the model demonstrates evasiveness regarding relevant issues, failing to provide clear responses.

\begin{table*}[!t]
\caption{Results of Generation Tasks}
\centering
\small
\begin{tabular}{lccc}
\toprule
\textbf{Method} & \textbf{Legal Text Summarization} & \textbf{Law Article Question Answering} & \textbf{Judicial Decision Reasoning} \\
\midrule
GPT-3.5 & 67.3 & 38.0 & 42.7 \\
Baichuan+SFT & 47.9 & 44.0 & 29.2 \\
\midrule
LuWen & \textbf{71.6} & \textbf{84.0} & \textbf{53.7} \\
\bottomrule
\end{tabular}
\label{tab: generate_res}
\end{table*}

\paragraph{Results of Generation Tasks}

From Table~\ref{tab: generate_res}, we have the following observations:
(1) Compared to the prediction tasks, generative language models demonstrated better fluency in the generation tasks, yielding relatively improved results.
(2) LuWen outperformed Baichuan+SFT across all tasks, which confirms that continual pre-training can significantly enhance a model's capability in handling law-related tasks.
(3) LuWen also displays a noticeable advantage over GPT-3.5, particularly in tasks such as law article question answering, which demand a high level of legal knowledge. This affirms the effectiveness of our technical approach.

\begin{figure*}[!t]
  \centering
  \includegraphics[width=\textwidth]{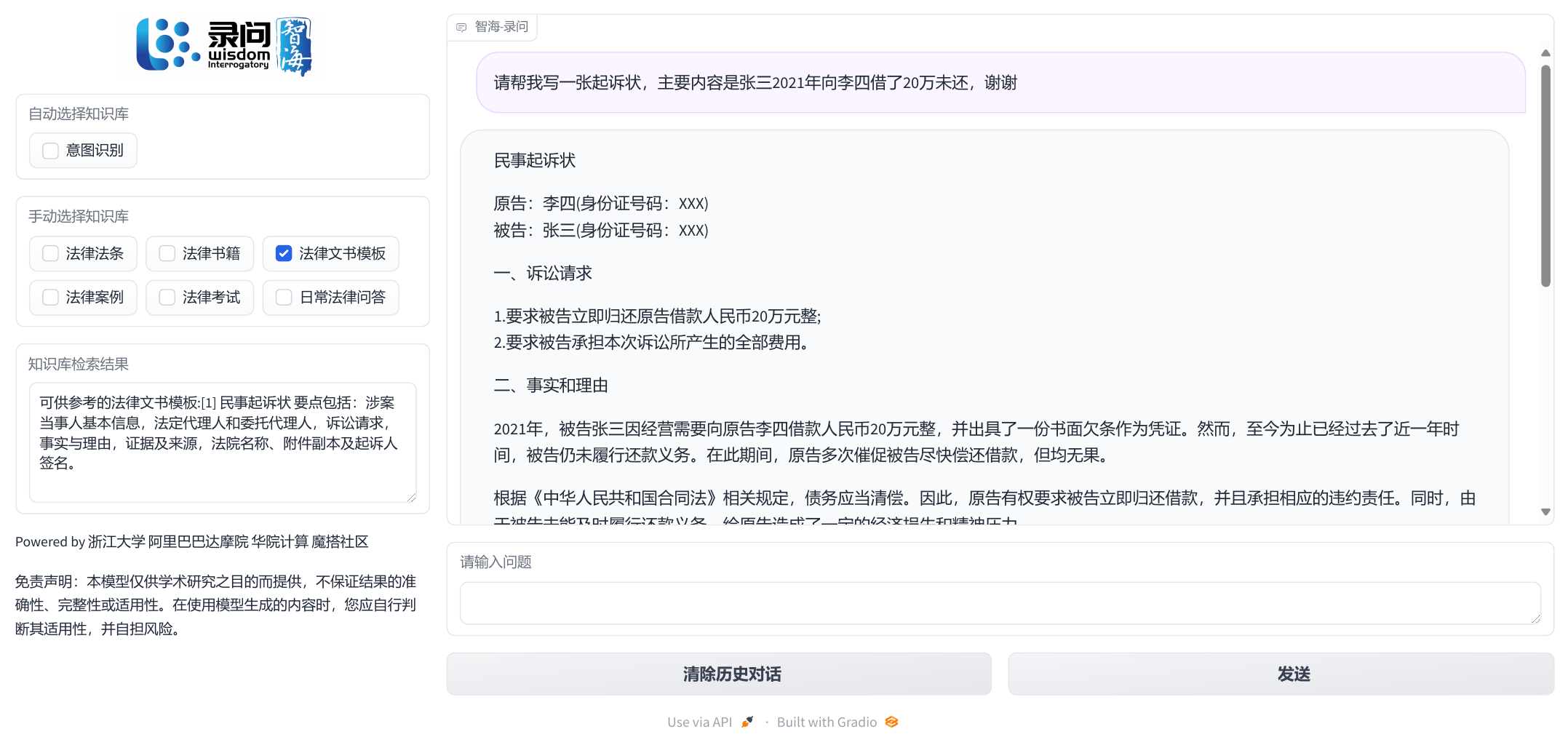}
  \caption{Sample Display of LuWen Added to the Retrieval Database}
  \label{fig:example-2}
\end{figure*}

In addition, Figure~\ref{fig:example-2} illustrates an example of how LuWen utilizes a retrieval database to assist in responding to user inputs. It demonstrates that LuWen can provide reasonable answers by integrating the given knowledge.

\section{Future Prospects}

Looking ahead, we aim to focus on the following directions to further deepen and expand the research and application scope of LuWen:

\paragraph{Task Decomposition Capability} Due to the inherent complexity of legal tasks, a single-step question-and-answer approach often fails to cover the necessary scope of solutions. Therefore, exploring methods for the model to autonomously and effectively decompose legal tasks becomes crucial for enhancing model performance. Integrating technologies such as agents \citep{wu2023autogen} or Chain-of-Thought (CoT) \citep{wei2022chain} represents a viable technical approach.

\paragraph{Retrieval Capability} Currently, retrieval based on vector and keyword matching remains inadequate. We plan to design a system that aims to automatically train retrieval models specifically tailored to particular knowledge bases, thereby improving the accuracy and efficiency of retrieval. Additionally, we acknowledge the model's limitations in understanding and integrating knowledge, and plan to enhance its capability in synthesizing and outputting integrated knowledge in future model training processes.

\paragraph{Multimodal Capability} Although most legal data exists in text form, the application scenarios in the legal field extend beyond text to include audio recordings, images, videos, and other data formats. Therefore, developing the model's multimodal capability is of critical importance to meet the complex requirements of the legal domain.

\balance
\bibliography{custom}

\end{document}